\documentclass[acmsmall,screen]{acmart}

\AtBeginDocument{%
  }

\setcopyright{acmcopyright}
\copyrightyear{2024}
\acmYear{2024}
\acmDOI{XXXXXXX.XXXXXXX}

\acmJournal{JACM}
\acmVolume{37}
\acmNumber{4}
\acmArticle{111}
\acmMonth{8}

\usepackage{algorithm}
\usepackage{algorithmic}
\usepackage{graphicx}
\usepackage{amsmath}
\usepackage{bm}
\usepackage{multirow}
\usepackage{enumitem}
\usepackage[switch]{lineno} 
\usepackage{animate}
\usepackage{amsmath,amsfonts}
\usepackage{array}
\usepackage[caption=false,font=normalsize,labelfont=sf,textfont=sf]{subfig}
\usepackage{textcomp}
\usepackage{stfloats}
\usepackage{url}
\usepackage{verbatim}
\usepackage{graphicx}
\usepackage{bm}
\usepackage{booktabs}
\usepackage{color}

\begin{document}
	
\title{SwinShadow: Shifted Window for Ambiguous Adjacent Shadow Detection}

\author{Yonghui Wang}
\email{wyh1998@mail.ustc.edu.cn}
\affiliation{%
  \institution{University of Science and Technology of China}
  \city{Hefei}
  \state{AnHui}
  \country{China}
  \postcode{230027}
}

\author{Shaokai Liu}
\email{liushaokai@mail.ustc.edu.cn}
\affiliation{%
	\institution{University of Science and Technology of China}
	\city{Hefei}
	\state{AnHui}
	\country{China}
	\postcode{230027}
}

\author{Li Li}
\email{lil1@ustc.edu.cn}
\affiliation{%
	\institution{University of Science and Technology of China}
	\city{Hefei}
	\state{AnHui}
	\country{China}
	\postcode{230027}
}

\author{Wengang Zhou}
\authornote{Corresponding author}
\email{zhwg@ustc.edu.cn}
\affiliation{%
	\institution{University of Science and Technology of China}
	\city{Hefei}
	\state{AnHui}
	\country{China}
}

\author{Houqiang Li}
\authornotemark[1]
\email{lihq@ustc.edu.cn}
\affiliation{%
	\institution{University of Science and Technology of China}
	\city{Hefei}
	\state{AnHui}
	\country{China}}

\renewcommand{\shortauthors}{Yonghui Wang et al.}

\begin{abstract}
Shadow detection is a fundamental and challenging task in many computer vision applications.
Intuitively, most shadows come from the occlusion of light by the object itself, resulting in the object and its shadow being contiguous (referred to as the adjacent shadow in this paper).
In this case, when the color of the object is similar to that of the shadow, existing methods struggle to achieve accurate detection.
To address this problem, we present SwinShadow, a transformer-based architecture that fully utilizes the powerful shifted window mechanism for detecting adjacent shadows.
The mechanism operates in two steps.
Initially, it applies local self-attention within a single window, enabling the network to focus on local details.
Subsequently, it shifts the attention windows to facilitate inter-window attention, enabling the capture of a broader range of adjacent information.
These combined steps significantly improve the network's capacity to distinguish shadows from nearby objects.
And the whole process can be divided into three parts: encoder, decoder, and feature integration.
During encoding, we adopt Swin Transformer to acquire hierarchical features.
Then during decoding, for shallow layers,  we propose a deep supervision (DS) module to suppress the false positives and boost the representation capability of shadow features for subsequent processing, while for deep layers, we leverage a double attention (DA) module to integrate local and shifted window in one stage to achieve a larger receptive field and enhance the continuity of information.
Ultimately, a new multi-level aggregation (MLA) mechanism is applied to fuse the decoded features for mask prediction.
Extensive experiments on three shadow detection benchmark datasets, SBU, UCF, and ISTD, demonstrate that our network achieves good performance in terms of balance error rate (BER).
The source code and results are now publicly available at \url{https://github.com/harrytea/SwinShadow}.
\end{abstract}


\received{20 February 2007}
\received[revised]{12 March 2009}
\received[accepted]{5 June 2009}

\maketitle
\section{Introduction}
\label{sec:intro}
Shadow appears whenever a light source is blocked partially or totally by an object during propagation. 
Knowing the shape, position, and intensity of shadows will provide us rich information such as the direction of light~\cite{lalonde2012estimating, panagopoulos2009robust} and the geometry of occluded objects~\cite{karsch2011rendering,okabe2009attached,moncrieff2008dynamic}.
Besides, the existence of shadow may impede various computer vision tasks, \emph{e.g.}, object detection~\cite{mikic2000moving,yu2018skeprid}, object tracking~\cite{cucchiara2001improving,guo2017efficient}, and semantic segmentation~\cite{finlayson2009entropy,singh2017secure}.
Hence, shadow detection has become a crucial task and attracted substantial research interest.

\begin{figure}[t]
	\centering
	\includegraphics[width=0.7\linewidth]{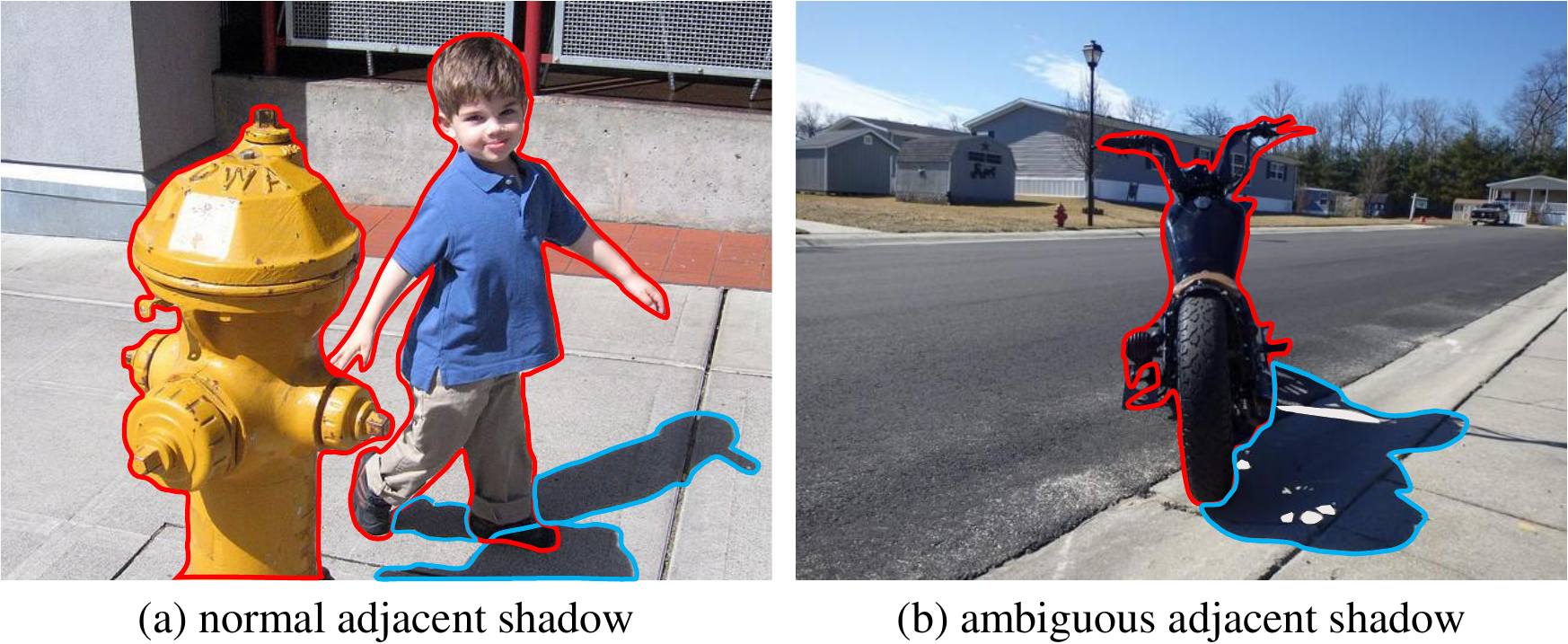}
	\caption{(a) and (b) are adjacent shadows that the object and its shadow are contiguous. The adjacent shadow in (a) has high contrast while (b) has lower contrast, which we refer to these two situation as normal adjacent shadow and ambiguous adjacent shadow, respectively.}
	\label{fig:def_adj}
\end{figure}

\begin{figure}[t]
	\centering
	\includegraphics[width=0.7\linewidth]{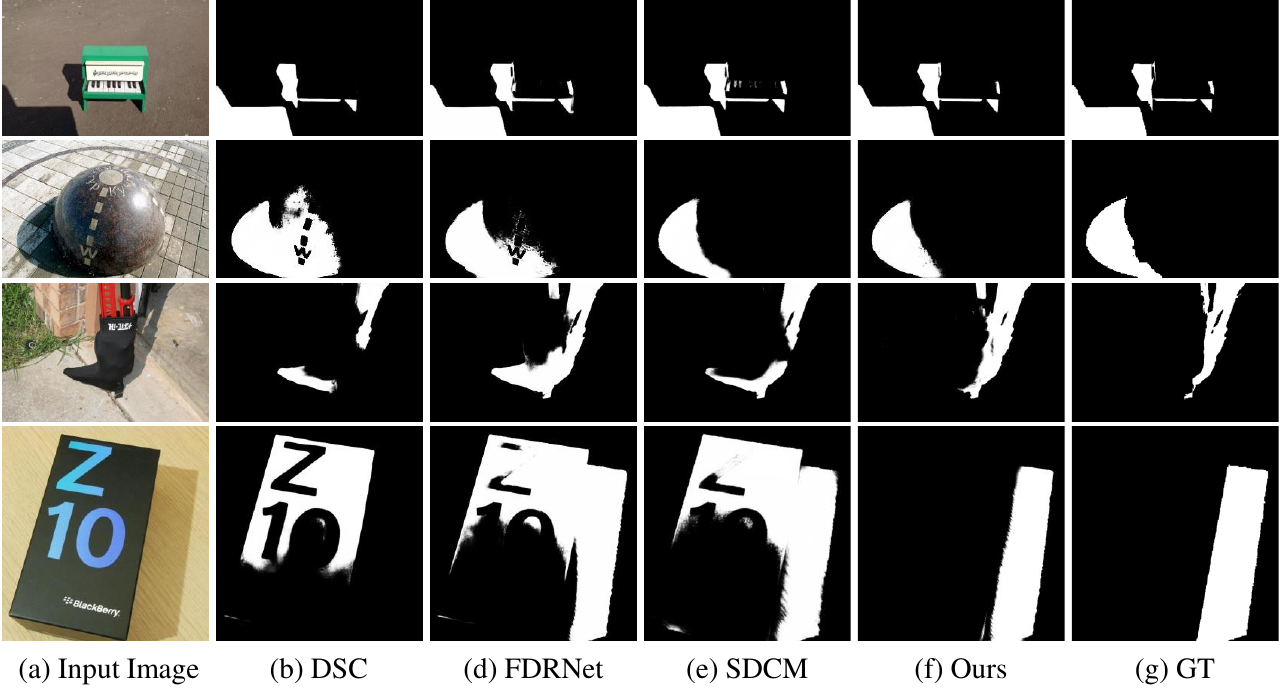}
	\caption{Adjacent shadows in shadow detection. Compared to row 1, the objects in rows 2, 3, and 4 have lower contrast with their shadows. As such, a more complex situation is that the color of the objects are darker than their shadows. Therefore, lacking attention on the objects and their adjacent regions will lead to mistaken shadow detection results.}
	\label{fig:intro}
\end{figure}

Previous works were proposed to detect shadows using physical models of color chromaticity or illumination~\cite{finlayson2009entropy,finlayson2005removal}.
Additionally, other works utilize hand-crafted features~\cite{zhang2007moving,huang2011characterizes,lalonde2010detecting,zhu2010learning} to estimate the shadow location.
However, the above methods may fail in natural scenes since the environment in the real world is very complex, and the model is challenging to produce a marked effect.
Recently, with the application of deep learning in computer vision tasks, data-driven methods have been explored in shadow detection tasks using CNNs~\cite{niu2022boundary,hu2018direction,zheng2019distraction,le2018a+,zhu2018bidirectional,chen2020multi,zhu2021mitigating,hu2019direction} and achieve superior detection results compared with the traditional methods.
Most CNN-based methods consider understanding global context information.
Such global semantics can well identify the cases in which the object and its shadow have high contrast (see Figure \ref{fig:def_adj} (a)).
As a consequence, recent methods~\cite{hu2018direction, chen2020multi, zhu2021mitigating} can detect these shadows well. 
While for more complex cases (see Figure \ref{fig:def_adj} (b)), where objects are darker than their shadows, they can not accurately detect.
For the above two situations, we call them \emph{normal adjacent shadow} and \emph{ambiguous adjacent shadow}, respectively.
Figure \ref{fig:intro} shows the comparison results of our method and other shadow detectors in adjacent shadow detection.
We argue that ignoring the delicate local information will lead to a limited detection effect on adjacent shadow, especially ambiguous adjacent shadow.
So we consider imposing local self-attention to alleviate the insufficient attention of local regions. 

\begin{figure}[t]
	\centering
	\includegraphics[width=0.7\linewidth]{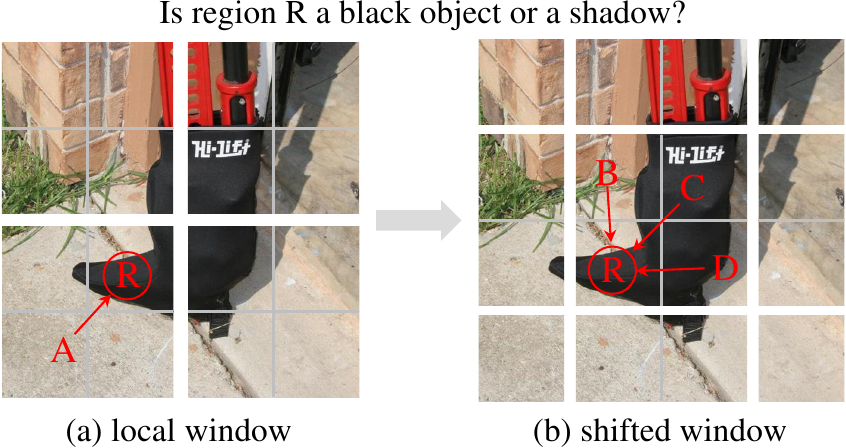}
	\caption{(a) Local window. Region R is a part of the black object and region A would give a strong indication that R is a shadow. (b) Shifted window. After shifting the window, region R can acquire more adjacent information (B, C, and D), giving R different cues that the properties of R and C (another part of the black object) are the same and R is a non-shadow region.}
	\label{fig:intro2}
\end{figure}

However, only adopting local window will lead to another problem: the network will ignore the information around the window, which is crucial for the detection of adjacent shadows.
Intuitively, the whole object usually occupies a part or a small part of the image.
Hence, using the local and shifted window can well understand the object and its surroundings.
In Figure~\ref{fig:intro2}, regions A and B represent the background, regions C and R are parts of the black object, and region D is the shadow cast by the black object.
As shown in Figure~\ref{fig:intro2}(a), for the window in the lower-left corner, region A would give a strong indication that compared to A, R is a shadow, leading to false recognition of the shadow. 
We point out that the key to solving the problem of ambiguous adjacent shadows is focusing on their surroundings.
Therefore, we introduce shifted window~\cite{liu2021swin} to acquire more adjacent information near the object. 
As shown in Figure \ref{fig:intro2}(b), after performing the local attention, we shift the attention window to obtain the sight of the surroundings.
Then, region R can acquire more adjacent information: region B, region C, and region D, where B and A are similar, C and R belong to a different part of the object, and D contains the shadow information of R.
Hence, based on the additional information, the random uncertainty of the model on region R will be eliminated~\cite{shannon1948mathematical}, and the black region R can be judged as a non-shadow region more accurately.

In this paper, we present a new transformer-based framework SwinShadow, which mainly solves the problem of ambiguous adjacent shadows by combining the local and shifted windows.
Our network is an encoder-decoder structure followed by a fusion mechanism.
We adopt swin transformer~\cite{liu2021swin} to encode features of different scales.
Swin Transformer~\cite{liu2021swin} is particularly suited to ambiguous adjacent shadow tasks.
It conducts the attention in the local window, allowing the model to focus on specific areas.
Additionally, its shifted window mechanism improves connectivity between adjacent windows, enhancing the model’s ability to distinguish shadows from nearby objects.
Moreover, the hierarchical structure of the Swin Transformer~\cite{liu2021swin} facilitates the model to assimilate high-level information that enriches its semantic understanding of the image.
This hierarchical architecture also supports pixel-level dense prediction tasks and enables the development of advanced feature fusion techniques.
Overall, these attributes collectively improve the model’s accuracy and efficiency in detecting adjacent shadows in complex visual scenes.
For the decoder, we first propose a deep supervision (DS) module embedded at the beginning of the network.
Providing this supervision can suppress non-shadow features and enhance shadow features to serve the subsequent processing.
Then, for the features of other scales, we adopt the double attention (DA) mechanism to conduct local and shifted attention simultaneously in one stage to strengthen the circulation of information.
Finally, we design an efficient feature fusion mechanism, multi-level aggregation (MLA) mechanism, to fuse the features of different scales and outputs, and then fuse these outputs to obtain the final score map.
We argue that the designed network can well resolve the shadow detection problem, especially ambiguous adjacent shadows.
Extensive experiments demonstrate that our method outperforms state-of-the-art methods over all three public benchmark datasets~\cite{vicente2016large,wang2018stacked,zhu2010learning}.

%
%
%

\section{Related Work}
\label{sec:work}
Shadow detection and removal are two main tasks in shadow fields and both of them have been investigated for many years~\cite{finlayson2005removal,benedek2008bayesian,finlayson2009entropy,tian2016new,vasluianu2023ntire}.
In this paper, we focus on shadow detection task.
In this section, we introduce previous works on shadow detection, including traditional and deep learning-based methods. Then we briefly review the self-attention mechanism, which is a crucial component in the transformer.

\noindent \textbf{Traditional methods.}
Early works mainly solve shadow detection by applying physical models, which identify shadow regions using the information of illumination~\cite{finlayson2005removal} and color~\cite{finlayson2009entropy,tian2009tricolor,amato2011accurate,tian2016new}.
Later, machine learning classifiers with hand-crafted features are proposed to further improve the accuracy of shadow detection~\cite{gevers2003classifying,plasberg2009feature,jung2009efficient,zhu2010learning,lalonde2010detecting,guo2011single,huang2011characterizes,vicente2015leave}.
Nevertheless, these designs only work well on high-quality images~\cite{lalonde2010detecting,nguyen2017shadow}. 
They often fail in complex real-world scenes since hand-crafted features have limited representation capability in describing shadows in the real world.

\noindent \textbf{Deep learning based methods.}
Several studies suggest that deep learning-based methods are beneficial to shadow detection task.
Khan \emph{et al.}~\cite{khan2014automatic} propose multiple CNNs to extract deep features from superpixels and target object boundaries for accurate shadow detection.
Then they feed the output features to a conditional random field (CRF) to further enhance the results.
Shen \emph{et al.}~\cite{shen2015shadow} employ local structures of shadow edges and formulate a structured CNN learning framework to improve the local consistency of the predicted shadow score map.
Vicente \emph{et al.}~\cite{vicente2016large} detect shadows using a stacked-CNN on a large dataset with noisy annotations.
Later, Nguyen \emph{et al.}~\cite{nguyen2017shadow} develop a generative adversarial network scGAN for shadow detection, while Hosseninzadeh \emph{et al.}~\cite{hosseinzadeh2018fast} use a patch-level CNN and a shadow prior map for fast shadow detection.

Very recently, Hu \emph{et al.}~\cite{hu2018direction} propose to detect shadows by learning global semantic context in a direction-aware manner.
Mohajerani \emph{et al.}~\cite{mohajerani2019shadow} apply a CNN-based network to preserve the image context and identify the global and local shadow attributes for shadow detection.
Le \emph{et al.}~\cite{le2018a+} combine a shadow detection network (D-Net) with a shadow attenuation network (A-Net) to detect the shadow regions.
Zhu \emph{et al.}~\cite{zhu2018bidirectional} design a recurrent attention residual module to combine global context and local context for shadow detection.
Wang \emph{et al.}~\cite{wang2018stacked} propose a GAN-based network to learn shadow detection and removal jointly.
Zheng \emph{et al.}~\cite{zheng2019distraction} propose to use distraction-aware features and design a novel structure for shadow detection.
Chen \emph{et al.}~\cite{chen2020multi} present a multi-task mean teacher model for semi-supervised shadow detection.
Zhu \emph{et al.}~\cite{zhu2021mitigating} introduce the concept of intensity bias to shadow detection and achieve superior shadow detection results.
Hu \emph{et al.}~\cite{hu2021revisiting} revisit the shadow detection problem, and present a more complex shadow detection dataset the future study.
Fang \emph{et al.}~\cite{fang2021robust} introduces a Effective-Context Augmentation (ECA) module to separate shadows from non-shadow objects.
Moreover, other researchers have also made additional explorations in this field~\cite{zhang2023exploiting,valanarasu2023fine,yang2023silt}.

Shadow detection performance is gradually improving on the benchmark datasets~\cite{vicente2016large,wang2018stacked,zhu2010learning}.
Most works focus on understanding global semantic information or properties of the image itself to enhance the model capacity.
However, the attention of the object and its surroundings is ignored, leading to unfriendly detection of adjacent shadows.

\noindent \textbf{Transformer.}
Transformer and self-attention modules have achieved great success in machine translation and NLP~\cite{dai2019transformer,kenton2019bert,vaswani2017attention,yang2019xlnet}.
Recently, such attention mechanism has shown beneficial in modeling high correlation in visual signals~\cite{hubel1962receptive,lecun1999object} and has started to be widely used in computer vision tasks.
The seminal work ViT~\cite{dosovitskiy2020image} proposes a pure transformer-based framework and has a dominant position in computer vision tasks~\cite{carion2020end,zhu2020deformable,zheng2021rethinking,wang2021end,chen2021pre}.
However, the self-attention mechanism brings quadratic computational complexity.
Some works tackle this problem by performing self-attention in non-overlapping local windows~\cite{liu2021swin,wang2022uformer,dong2022cswin}, which achieve linear complexity and significantly reduce the computational overhead.

In the shadow detection task, existing methods still misrecognize some black objects in ambiguous adjacent shadows.
Intuitively, these objects and shadows only account for a part of the image, so local attention can well focus on them.
Furthermore, only performing local attention is insufficient to understand the surroundings of the object.
Therefore, we introduce the shifted window mechanism to solve this problem. 
We believe that local attention with shifted window mechanism can provide adaptive attention and solve the ambiguous adjacent shadows well. 
Results show that our method can further outperform in terms of BER value on all three benchmark datasets~\cite{vicente2016large,wang2018stacked,zhu2010learning}.

\begin{figure*}[t]
	\centering
	\includegraphics[width=1.0\linewidth]{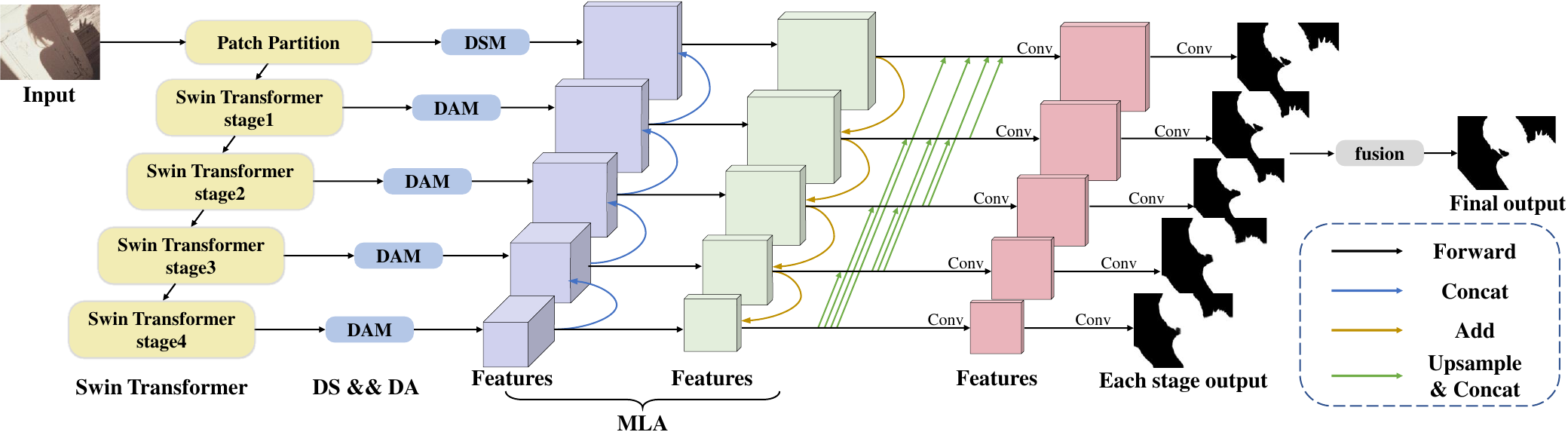}
	\caption{Schematic illustration of the proposed SwinShadow. We first split an image into fixed-size patches and linearly embed each of them. Then we feed the sequence of vectors to a Swin Transformer encoder~\cite{liu2021swin}. To obtain accurate shadow detection results, we use the deep supervision (DS) module to process patch partition features and double attention (DA) module to process high-level features. Last, a multi-level feature aggregation (MLA) mechanism is applied to fuse the features, and we use these features for the final output.}
	\label{fig:arch}
\end{figure*}

\section{Methodology}
\label{sec:method}

\begin{figure}[t]
	\centering
	\includegraphics[width=0.7\linewidth]{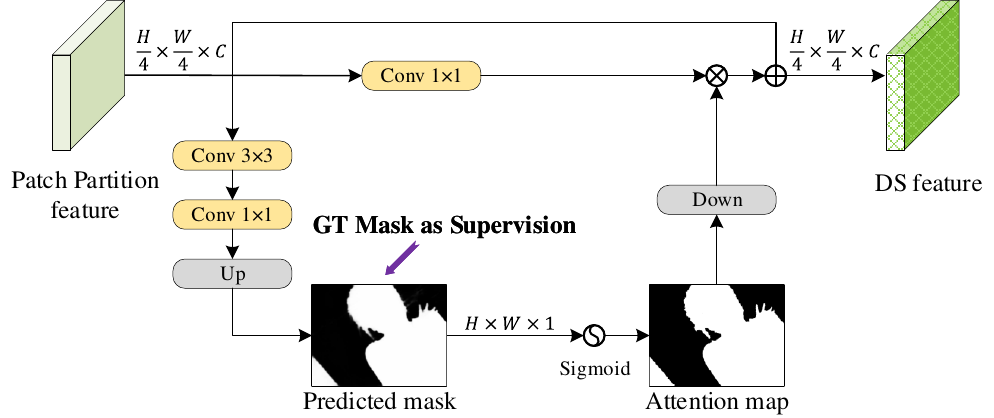}
	\caption{Deep supervision module. First, image features from the patch partition module of the Swin Transformer encoder~\cite{liu2021swin} are processed by two convolutions followed by an upsampling operation to obtain predicted map, and we add supervision to this map. Then the predicted map is passed to a sigmoid activation function to obtain attention map. Finally, we use this attention map to enhance our features and acquire deep supervision (DS) feature through a skip connection.}
	\label{fig:dsm}
\end{figure}

Figure \ref{fig:arch} illustrates the overall architecture of the proposed SwinShadow that employs shifted window mechanism to focus on local and adjacent regions.
The network takes the whole image as input and processes it in an end-to-end manner.
First, the transformer-based encoder is applied to extract a series of hierarchical features containing fine details and rich semantic information over different scales.
Second, for the feature which has high resolutions, we design a deep supervision (DS) module (see Figure \ref{fig:dsm}) to enhance shadow signals and suppress the non-shadow signals over all regions. 
Such a supervision mechanism can assist in optimizing feature extraction at the beginning of the network, benefiting all subsequent modules rather than only relying on the supervision at the end of the network. 
Moreover, we utilize the double attention (DA) module to decode adjacent shadow signals of high-level features encoded by the swin transformer.
Third, we propose a multi-level feature aggregation (MLA) mechanism to refine feature maps and utilize the improved features to predict score maps at each layer. 
Finally, we apply a convolution layer (via a $1\times 1$ kernel) to fuse all predicted score maps and obtain the final output. 
The whole process is described next in detail.

\subsection{Transformer-based Encoder}
As we described in Section \ref{sec:intro}, for the detection of adjacent shadows, local attention is crucial to focus on the partial area of the image, \emph{e.g.}, a black object or its shadow.
Additionally, we need more cues surrounding the object to precisely judge the ambiguous adjacent shadow.
Therefore, an extra shifted window mechanism is introduced, which we can obtain the other information around the object, such as other parts of itself, its shadow, and its surroundings.

Swin transformer~\cite{liu2021swin}, designed in hierarchical architecture and computing the representation signals with local and shifted windows, is a general-purpose feature extraction backbone in numerous computer vision tasks.
It is divided into four stages, with each stage consisting of several Swin Transformer blocks. 
These stages proceed sequentially, gradually refining the feature representation, for instance, reducing the resolution while increasing the channel dimension of feature maps.
Each stage processes features at different scales, thereby obtaining multi-level features for subsequent feature processing.
The design of local windows can limit the calculation of self-attention to the interior of windows, and there is non-overlapping between windows, reducing the computation costs.
Moreover, the proposed shifted windows can allow flexible interaction between adjacent windows, which meets our requests well. 
Specifically, for two consecutive swin transformer blocks, given an input feature map $\bm{x}^{l}\in \mathbb{R}^{H \times W \times C}$, the whole operation can be described as follows:
\begin{equation}
	\begin{aligned}
		& \bm{\hat{x}}^{l+1} = \operatorname{W-MSA}(\operatorname{LN}(\bm{x}^{l})) + \bm{x}^{l}, \\
		& \bm{x}^{l+1} = \operatorname{MLP}(\operatorname{LN}(\bm{\hat{x}}^{l+1})) + \bm{\hat{x}}^{l+1}, \\
		& \bm{\hat{x}}^{l+2} = \operatorname{SW-MSA}(LN(\bm{x}^{l+1})) + \bm{x}^{l+1}, \\
		& \bm{x}^{l+2} = \operatorname{MLP}(\operatorname{LN}(\bm{\hat{x}}^{l+2})) + \bm{\hat{x}}^{l+2},
	\end{aligned}
	\label{eq:wmsa}
\end{equation}
where $\bm{\hat{x}}^{l+1}$ and $\bm{x}^{l+1}$ denote the output feature map of $\operatorname{(S)W-MSA}$ and $\operatorname{MLP}$ module for block $l$, respectively. 
$\operatorname{W-MSA}$ and $\operatorname{SW-MSA}$ denote window-based multi-head self-attention under the regular and shifted window partitioning settings, respectively.
Such window-based attention combined with interaction across adjacent windows can effectively boost the modeling capacity for adjacent shadow.

\subsection{Transformer-based Decoder}
Swin transformer produces hierarchical feature maps, and such multi-level representations provide diverse fine details and semantic visual concepts over different scales.
Here, our decoder is divided into two parts, the deep supervision (DS) module and the double attention (DA) module. 
The first part imposes supervision on the embedded features before entering the transformer, which enhances the shadow features and simultaneously suppresses the non-shadow features.
The second part decodes the features of swin transformer blocks and combines two consecutive blocks into one, which reduces the long-range dependence and effectively improves the shadow detection performance.
The two modules are described next.

\noindent \textbf{Deep supervision module.} 
Previous methods apply a deep supervision mechanism to impose deep supervision signals at the end of each layer. 
However, deep supervision in the middle process of features is less explored in shadow detection.
Inspired by the success of the supervision in computer vision tasks~\cite{zamir2021multi,lu2022video}, we design a simple supervision module to refine the features effectively.
Figure \ref{fig:dsm} shows the details of our proposed deep supervision module.
Compared with the supervision in~\cite{zamir2021multi}, the main difference is that the attention map in our module only has one channel, which can equally suppress (or enhance) all features while reducing computational overhead.
With the decoded features $\bm{F}_{in}\in \mathbb{R}^{\frac{H}{4} \times \frac{W}{4} \times C}$ as input, first the DS module uses a $3\times 3$ convolution and a $1\times 1$ convolution to process the incoming features and reduce the number of feature channels to 1. 
Then an upsampling operation is applied to obtain the predicted map $\bm{M}_{p}\in \mathbb{R}^{H \times W \times 1}$. 
For this score map, we provide direct supervision with the ground-truth map.
Next, the attention map $\bm{M}_{a}\in \mathbb{R}^{H \times W \times 1}$ is generated from the predicted map $\bm{M}_{p}$ by using a sigmoid activation function.
After a downsampling operation, the attention map $\bm{M}_{a}$ is applied to re-calibrate the transformed input features $\bm{\hat{F}}_{in}$, which is obtained by a $1\times 1$ convolution. Finally, these attention-guided features are added to the original input features, resulting in deep supervision (DS) features $\bm{F}_{out}$.

\noindent \textbf{Double attention module.}
For two consecutive swin transformer blocks, they perform  $\operatorname{W-MSA}$ and $\operatorname{SW-MSA}$, respectively.
However, such separate operations may lead to discontinuous information transfer and limit the receptive fields in the same number of blocks.
Similar to~\cite{zhang2022styleswin}, we apply \emph{double attention} mechanism to improve the above situation.
Starting from feature map $\bm{F}_{dim}\in \mathbb{R}^{H \times W \times C}$, we split the feature channels into two parts: $\bm{F}^{1}_{dim}\in \mathbb{R}^{H \times W \times \frac{C}{2}}$ and $\bm{F}^{2}_{dim}\in \mathbb{R}^{H \times W \times \frac{C}{2}}$. 
The first half performs regular window attention, while the second half performs shifted window attention.
After conducting attention separately, both of the processed features are concatenated to produce the output $\bm{\hat{F}}_{dim}$. 
The whole process can be formulated as follows:

\begin{equation}
	\begin{aligned}
		& \bm{F}_{dim}^{1}, \bm{F}_{dim}^{2} = \operatorname{Split}(\bm{F}_{dim}), \\
		& \bm{x}_{dim}^{1}, \bm{x}_{dim}^{2} = \operatorname{Window}(\bm{F}_{dim}^{1}), \operatorname{Shifted}(\bm{F}_{dim}^{2}), \\
		& \bm{\hat{F}}_{dim}^{1}, \bm{\hat{F}}_{dim}^{2} = \operatorname{Attn}(\bm{x}_{dim}^{1}), \operatorname{Attn}(\bm{x}_{dim}^{2}), \\
		& \bm{\hat{F}}_{dim} = \operatorname{Concat}(\bm{\hat{F}}_{dim}^{1}, \bm{\hat{F}}_{dim}^{2}),
	\end{aligned}
	\label{eq:dam}
\end{equation}
where $\bm{x}_{dim}^{1}$ and $\bm{x}_{dim}^{2}$ are non-overlapping patches under regular and shifted window partitioning settings, respectively. $\operatorname{Attn}$ denotes window-based multi-head self-attention. 
\emph{Double attention} module can capture the local and adjacent information in a single block, which reduces the long-range dependency and enlarge the receptive field, resulting in adaptive adjacent shadow detection.

\begin{figure}[t]
	\centering
	\includegraphics[width=0.7\linewidth]{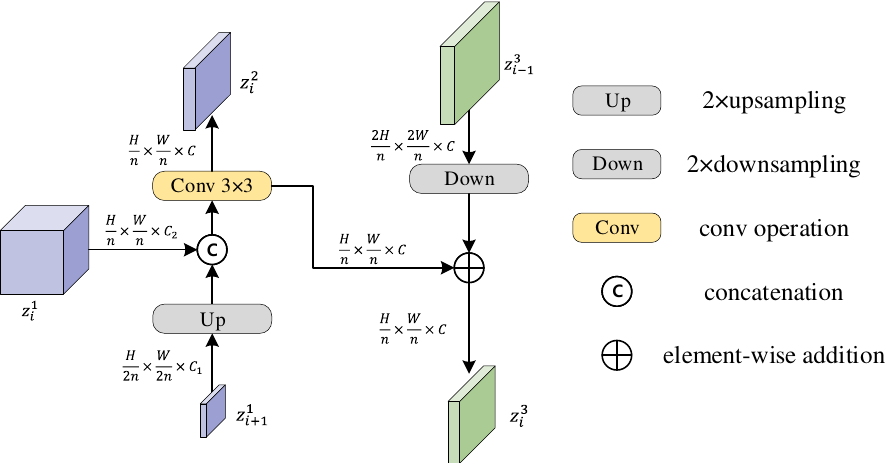}
	\caption{Multi-level aggregation mechanism. The decoded feature $z_{i}^{1}$ is first refined through $z_{i+1}^{1}$ to obtain $z_{i}^{2}$. Then we downsample the feature $z_{i-1}^{3}$ and add it to $z_{i}^{2}$ to acquire the fusion feature $z_{i}^{3}$.}
	\label{fig:fusion}
\end{figure}

\subsection{Multi-level feature Aggregation}
Most existing shadow detection methods fuse the multi-scale feature maps using short connections proposed in DSS~\cite{hou2017deeply}. 
In our observation, such a connection has two limits.
First, the channel of high-level features is much higher than that of the original input features. 
Simply fusing them will introduce high computation costs and make the feature information challenging to utilize and propagate.
Second, the difference of crucial subject information between different scales varies wildly, \emph{e.g.}, more details in high resolution features and more semantics in high-level features.
To solve the problem above, we propose a feature fusion method based on FPN~\cite{lin2017feature}.
In our experiments, we find that the feature with high resolution is crucial to the details and localization accuracy of shadow masks.
Therefore, another bottom-up aggregation structure is applied.
As shown in Figure \ref{fig:arch} and Figure \ref{fig:fusion}.
Each decoded feature map $\bm{z}_{i}^{1}$ first concatenates with an up-sampled high-level feature $\bm{z}_{i+1}^{1}$, and then the feature goes through a $3\times 3$ convolution to reduce the channel size and produce the new feature $\bm{z}_{i}^{2}$.
Next, each feature map $\bm{z}_{i}^{2}$ and the down-sampled map $\bm{z}_{i-1}^{3}$ are added through lateral connection, where $i$ denotes the feature at different stage, and the number in the upper right corner of $z$ represents the times the feature has been processed after the decoder.

Finally, we use a short connection to fuse these feature maps and predict score maps at each level.
By doing this, features of different scales can interact well, and each scale can be refined through others.

\subsection{Training and Testing Strategies}
\textbf{Loss function.}
In natural scene images, the ratio of shadow and non-shadow pixels are usually imbalanced.
Considering a shadow image $\bm{I_{s}}$, we assume that the ground truth label for a pixel is $y_{i}$ ($y=1$ for shadow pixel and $y=0$ for non-shadow pixel) and the prediction label of the pixel is $p_{i}$ ($p_{i}\in [0, 1]$). To balance the weight between shadow pixels and non-shadow pixels, we adopt a weighted cross entropy loss as shadow detection loss:

\begin{equation}
	\begin{aligned}
		L = -\sum_{j}( \frac{N_{n}}{N_{n}+N_{p}}y_{i}\log(p_{i})+ \frac{N_{p}}{N_{n}+N_{p}}(1-y_{i})\log (1-p_{j})),
	\end{aligned}
	\label{eq:loss}
\end{equation}
where $N_{p}$ and $N_{n}$ are the number of shadow and non-shadow pixels, respectively.

\textbf{Training parameters.}
Our network is built on Swin-B~\cite{liu2021swin} as an encoder for extracting multi-level features. 
The Swin-B is pre-trained on the ImageNet~\cite{deng2009imagenet} to accelerate the training procedure and reduce the overfitting risk, whereas the other parameters are randomly initialized. 
Stochastic gradient descent~\cite{qian1999momentum}, with a momentum value of $0.9$ and a weight decay of $5\times 10^{-4}$ is used to optimize the whole network for $20k$ iterations.
We set the learning rate as $5\times 10^{-3}$.
Additionally, the training data is augmented by random horizontal flipping and then the image is resized to $384\times 384$.
We build the model based on PyTorch with a mini-batch size of 4 and the training is run on a single RTX2080Ti GPU.

\textbf{Inference.}
During testing, we resize the input image to the exact resolution as training does.
Then, the network produces one score map in each layer and fuses them with a $1\times 1$ convolution to obtain the output score map.
Finally, we apply the fully connected conditional field (CRF)~\cite{krahenbuhl2011efficient}, which is widely used in the shadow detection task, to further refine the predicted score map by optimizing the spatial coherence of each pixel and obtain the final score map.

\section{Experiments}
\label{exp}

\subsection{Datasets and Evaluation Metircs}
\label{exp:data}

\textbf{Benchmark datasets.}
We evaluate our methods on three widely-used shadow detection benchmark datasets: SBU~\cite{vicente2016large}, UCF~\cite{zhu2010learning}, and ISTD~\cite{wang2018stacked}. 
All of them have shadow images and their corresponding annotated shadow masks. 
Specifically, The SBU dataset is the largest publicly available annotated shadow detection dataset, containing $4089$ training images and $638$ testing images. 
The UCF dataset contains 135 training images and 110 testing images, while the ISTD dataset contains $1330$ training images and $540$ testing images. 
Following previous shadow detection methods, we train on the SBU training set and then test on the SBU testing set and the UCF testing set. 
For the evaluation of the ISTD dataset, we train on its training set.

\begin{figure*}[t]
	\centering
	\includegraphics[width=0.98\linewidth]{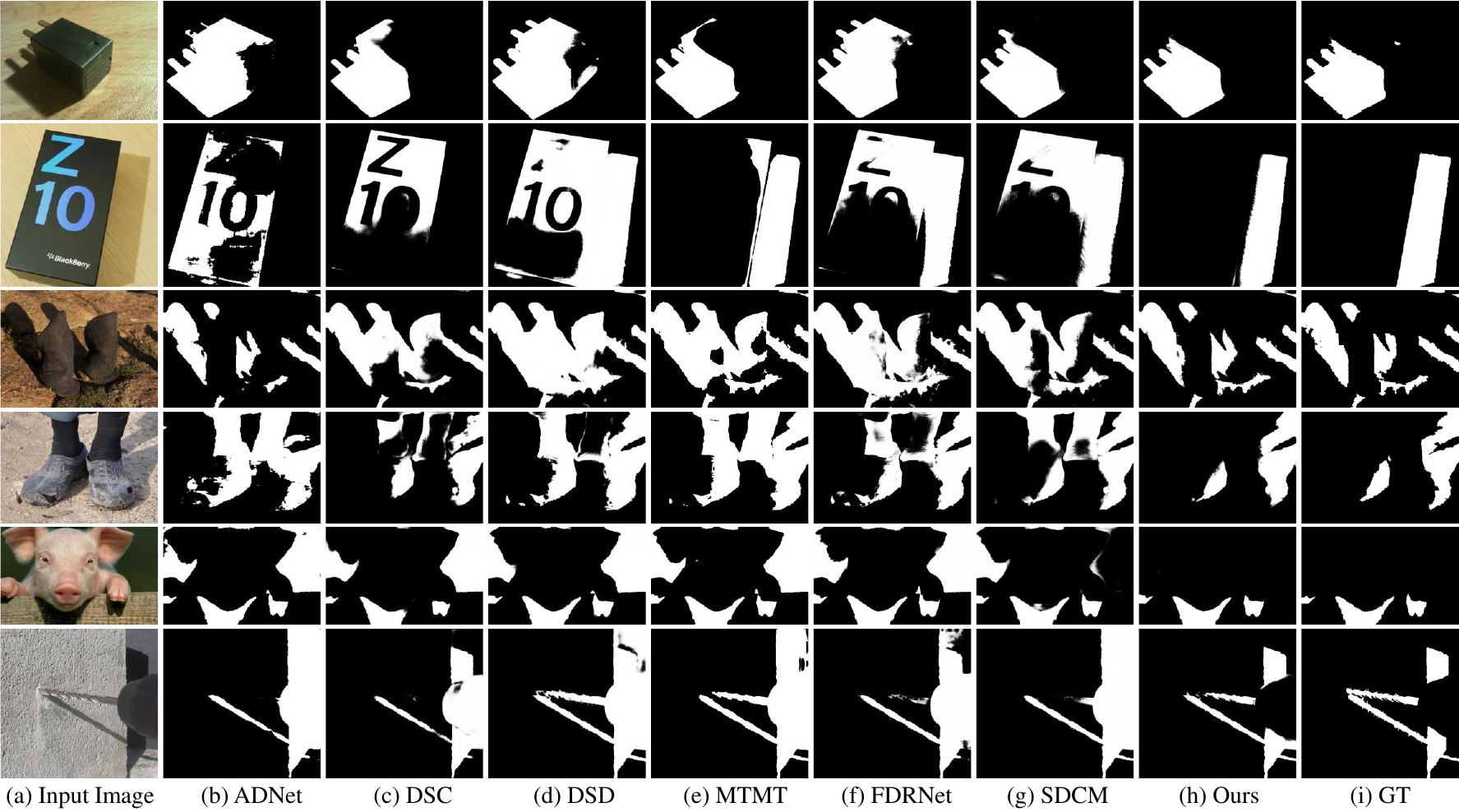}
	\caption{Qualitative comparison of our method with the state-of-the-art shadow detection methods. The first four rows show the results of ambiguous adjacent shadow detection, and the last two rows show another two challenging cases.}
	\label{fig:qualitative}
\end{figure*}

\begin{table*}[t]
	\renewcommand\arraystretch{1.05}  
	\caption{Quantitative comparison of our method with the state-of-the-art methods on three shadow detection benchmark datasets. ``BER'' represents the balance error rate considering both shadow and non-shadow regions. ``Shadow'' and ``Non Shad.'' denote the BER values for shadow and non-shadow regions, respectively. Best results are highlighted in bold. ``$\downarrow$'' indicates the lower the better.}
	\centering
	\scalebox{0.7}{
		\begin{tabular}{
				m{2.9cm}<{\centering}
				|m{0.65cm}<{\centering}
				|m{0.9cm}<{\centering}m{1cm}<{\centering}m{1.7cm}<{\centering}
				|m{0.9cm}<{\centering}m{1cm}<{\centering}m{1.7cm}<{\centering}    
				|m{0.9cm}<{\centering}m{1cm}<{\centering}m{1.7cm}<{\centering}
			}
			\hline    
			&  & \multicolumn{3}{c|}{SBU} & \multicolumn{3}{c|}{UCF} & \multicolumn{3}{c}{ISTD}  \\
			\hline
			methods & year & BER$\downarrow$ & Shadow$\downarrow$ & Non Shad.$\downarrow$ & BER$\downarrow$ &  Shadow$\downarrow$ & Non Shad.$\downarrow$ & BER$\downarrow$ & Shadow$\downarrow$ & Non Shad.$\downarrow$ \\
			\hline
			PSPNet~\cite{zhao2017pyramid}           & 2017 & 8.57  & -     & -        & 11.75 & -     & -        & 4.26  & 4.51  & 4.02 \\			
			SRM~\cite{wang2017stagewise}            & 2017 & 6.51  & 10.52 & 2.50     & 12.51 & 21.41 & 3.60     & 7.92  & 13.97 & 1.86 \\
			RAS~\cite{chen2018reverse}              & 2018 & 7.31  & 12.13 & 2.48     & 13.62 & 23.06 & 4.18     & 11.14 & 19.88 & 2.41 \\			
			EGNet~\cite{zhao2019egnet}              & 2019 & 4.49  & 5.23  & 3.75     & 9.20  & 11.28 & 7.12     & 1.85  & 1.75  & 1.95 \\			
			ITSD~\cite{zhou2020interactive}         & 2020 & 5.00  & 8.65  & 1.36     & 10.16 & 17.13 & \textbf{3.19}     & 2.73  & 2.05  & 3.40 \\
			\hline
			Unary-Pairwise~\cite{guo2011single}     & 2011 & 25.03 & 36.26 & 13.80    & -     & -     & -        & -     & -     & - \\
			scGAN~\cite{wang2018stacked}            & 2017 & 9.10  & 8.39  & 9.69     & 11.50 & 7.74  & 15.30    & 4.70  & 3.22  & 6.18 \\
			patched-CNN~\cite{hosseinzadeh2018fast} & 2018 & 11.56 & 15.60 & 7.52     & -     & -     & -        & -     & -     & - \\
			ST-CGAN~\cite{wang2018stacked}          & 2018 & 8.14  & 3.75  & 12.53    & 11.23 & \textbf{4.94}  & 17.52     & 3.85 & 2.14  & 5.55 \\
			BDRAR~\cite{zhu2018bidirectional}       & 2018 & 3.64  & 3.40  & 3.89     & 7.81  & 9.69  & 5.44     & 2.69  & \textbf{0.50}  & 4.87 \\
			DSC~\cite{hu2018direction}              & 2018 & 5.59  & 9.76  & \textbf{1.42}     & 10.38 & 11.72 & 3.04      & 3.42 & 3.85  & 3.00 \\
			ADNet~\cite{le2018a+}                   & 2018 & 5.37  & 4.45  & 6.30     & 9.25  & 8.37  & 10.14    & -     & -     & - \\
			DC-DSPF~\cite{wang2018densely}          & 2019 & 4.90  & 4.70  & 5.10     & 7.90  & 6.50  & 9.30     & -     & -     & - \\	
			DSD~\cite{zheng2019distraction}         & 2019 & 3.45  & 3.33  & 3.58     & 7.52  & 9.52  & 5.53     & 2.17  & 1.36  & 2.98 \\
			MTMT~\cite{chen2020multi}               & 2020 & 3.15  & 3.73  & 2.57     & 7.47  & 10.31 & 4.63     & 1.75 & 1.36  & 2.14 \\		
			FDRNet~\cite{zhu2021mitigating}         & 2021 & 3.04  & 2.91  & 3.18     & 7.28  & 8.31  & 6.26     & 1.55 & 1.22  & 1.88 \\
			
			RMLANet~\cite{jie2022rmlanet}         & 2022 & 2.97  & \textbf{2.53}  & 3.42     & 6.41  & 6.69  & 6.14     & \textbf{1.01} & 0.68  & \textbf{1.34} \\
			SDCM~\cite{zhu2022single}         & 2022 &  2.94 &  - &  -   & 6.73  &  8.15 &  5.31    & 1.44 & 1.19  & 1.69 \\
			R2D~\cite{valanarasu2023fine}         & 2023 &  3.15 & 2.74 &  3.56  & 6.96  &  8.32 &  5.60    & 1.69 & 0.59  & 2.79 \\
			\hline   
			\textbf{Ours} & - & \textbf{2.78} & 2.78 & 2.79    & \textbf{5.99} & 6.57 & 5.40    & 1.22 & 1.05 & 1.40 \\    
			\hline
		\end{tabular}
	}
	\label{tab:quantitative}
\end{table*}

\textbf{Evaluation Metrics.}
We employ balance error rate (BER), commonly used in the previous methods, to evaluate the quantitative results of our method. This metric can be computed as follows:    
\begin{equation}
	BER = (1-\frac{1}{2}(\frac{TP}{TP+FN}+\frac{TN}{TN+FP}))\times 100,
	\label{eq:ber}
\end{equation}
where TP,TN,FP and FN denote the number true positives, true negatives, false positives and false negatives, respectively. 
The lower the BER is, the better the performance will be.

\subsection{Comparison with the State-of-the-art Shadow Detectors}
\label{exp:detectors}

We compare our method with 14 recent state-of-the-art shadow detectors, including R2D~\cite{valanarasu2023fine}, SDCM~\cite{zhu2022single}, RMLANet~\cite{jie2022rmlanet}, FDRNet~\cite{zhu2021mitigating}, MTMT~\cite{chen2020multi}, DSD~\cite{zheng2019distraction}, DC-DSPF~\cite{wang2018densely}, ADNet~\cite{le2018a+}, DSC~\cite{hu2018direction}, BDRAR~\cite{zhu2018bidirectional}, ST-CGAN~\cite{wang2018stacked}, patched-CNN~\cite{hosseinzadeh2018fast}, scGAN~\cite{wang2018stacked}, and Unary-Pairwise~\cite{guo2011single}. 
Among them, only Unary-Pairwise is based on hand-crafted features, whereas others are deep learning-based methods.
For a fair comparison, we adopt the shadow detection results directly from the authors or use their public code to produce final results.

\begin{figure*}[t]
	\centering
	\includegraphics[width=0.98\linewidth]{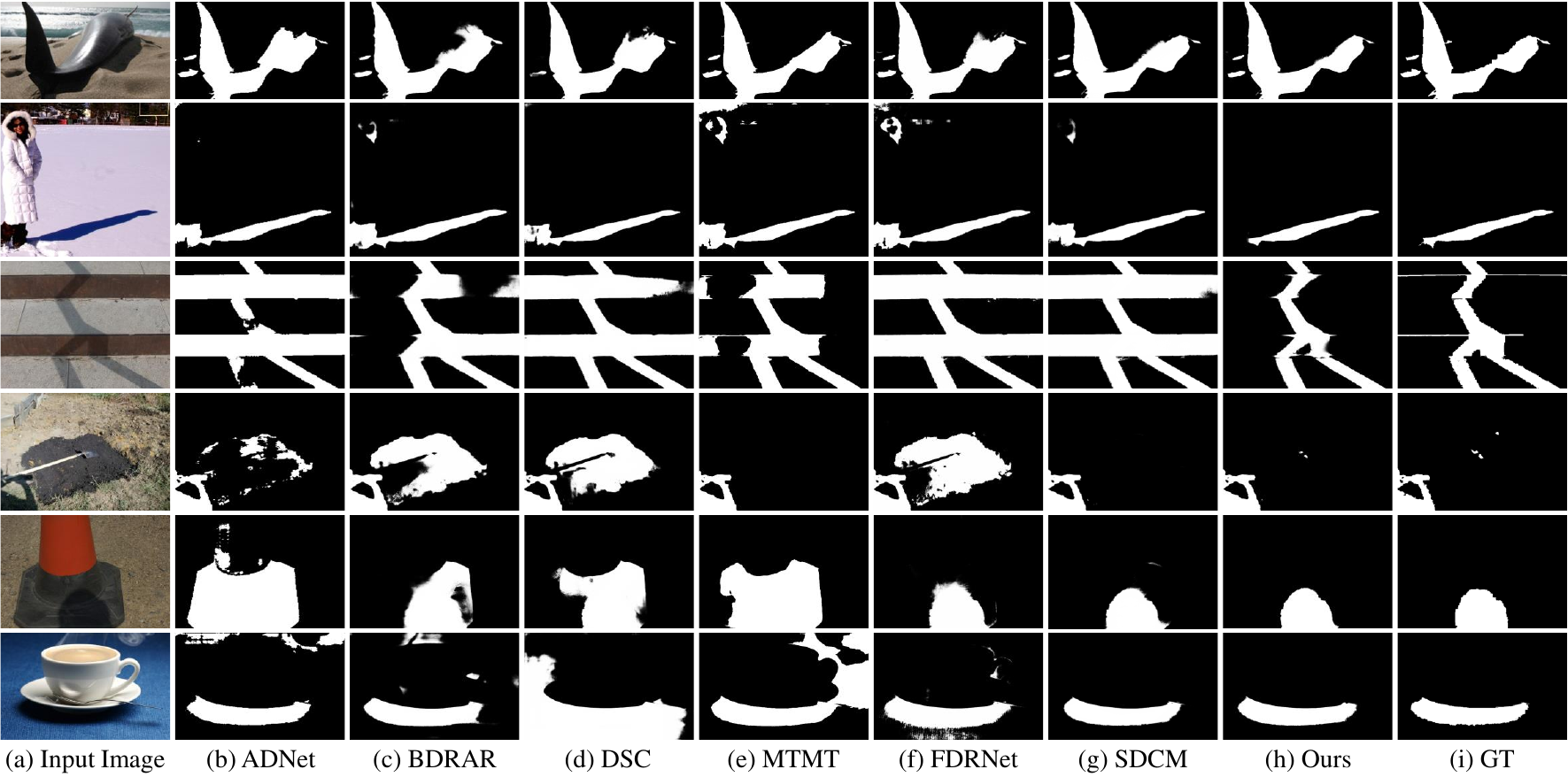}
	\caption{More qualitative comparison of our method with the state-of-the-art shadow detection methods.}
	\label{fig:add_qualitative}
\end{figure*}

Table \ref{tab:quantitative} shows the quantitative results of different methods on the three benchmark datasets.
The BER score represents the mean of the shadow BER score and non-shadow BER score.
Our method achieves the best BER results across the whole region on SBU~\cite{vicente2016large} and UCF~\cite{zhu2010learning} datasets.
Compared to the method FDRNet~\cite{zhu2021mitigating}, which adjusts the intensity bias to address the misrecognition of dark non-shadow regions and bright shadow regions, our method reduces the BER scores by $8.6\%$, $17.7\%$ and $21.3\%$ on SBU, UCF, and ISTD dataset, respectively.
Additionally, the BER score predicted by our method for shadow and non-shadow pixels is very close. 
In contrast, others prefer one of the two, indicating that our method is more balanced in detecting shadow and non-shadow regions and reduces false positive and false negative predictions.


To further evaluate the effectiveness of our method, we show some visual results to qualitatively compare with the state-of-the-art methods.
As shown in Figure \ref{fig:qualitative}, we present some cases from the top four rows which are ambiguous adjacent shadows.
The objects are similar to their shadows or even darker, making the model difficult to distinguish.
However, our method can solve such ambiguous cases well.
Furthermore, in the last two rows, we present another two challenging cases: a pig with a complex dark background and a black object around the shadow.
It is hard to identify whether it is a shadow without leveraging local self-attention on the object and its adjacent regions.
From these visual results, we can conclude that our method can effectively recognize adjacent shadows and avoid the false recognition of black non-shadow objects, making our method more applicable.
Figure \ref{fig:add_qualitative} shows more qualitative comparison results.

\begin{figure}[t]
	\centering
	\includegraphics[width=0.7\linewidth]{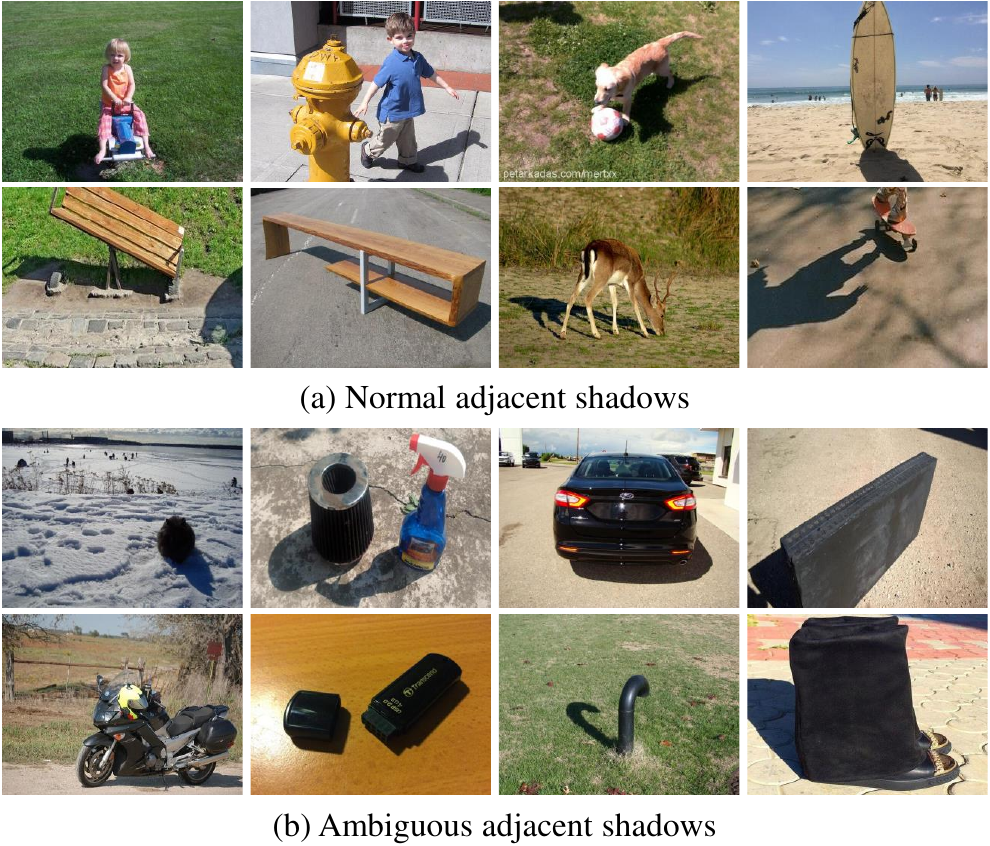}
	\caption{Some cases of normal adjacent shadows and ambiguous adjacent shadows in SBU~\cite{vicente2016large} dataset.}
	\label{fig:adj}
\end{figure}

\subsection{Comparison with Saliency Detection and Semantic Segmentation Methods}
\label{exp:segment}
Shadow detection is a pixel-level classification problem.
Deep learning networks designed for saliency detection and semantic segmentation can be re-trained on the shadow detection datasets and used for shadow detection.
To further evaluate the effectiveness of our method, we conduct another experiment with five saliency detection and semantic segmentation methods, \emph{i.e.}, ITSD~\cite{zhou2020interactive}, EGNet~\cite{zhao2019egnet}, RAS~\cite{chen2018reverse}, SRM~\cite{wang2017stagewise}, and PSPNet~\cite{zhao2017pyramid}.
We re-train these publicly available models and adjust the parameters for the best performances.
The first five rows of Table \ref{tab:quantitative} report the quantitative comparison results.
Even though they have better BER scores than some of the shadow detection methods, which indicates that the task of saliency detection and semantic segmentation are also applicable to shadow detection, our method still outperforms them for the three benchmarks.

\begin{table}[t]
	\renewcommand\arraystretch{1.1}  
	\caption{Percentage of adjacent shadows and ambiguous adjacent shadows in the SBU and UCF dataset. ``Adjacent'' represents the ratio of adjacent shadows in the whole dataset and ``Ambiguous'' represents the ratio of ambiguous adjacent shadows in the whole dataset.}
	\centering
	\scalebox{0.95}{
		\begin{tabular}{
				m{1cm}<{\centering}
				|m{1.5cm}<{\centering}m{1.5cm}<{\centering}
				|m{1.5cm}<{\centering}m{1.5cm}<{\centering}
			}
			\hline
			& \multicolumn{2}{c|}{Training set} & \multicolumn{2}{c}{Testing set}  \\
			& Adjacent & Ambiguous & Adjacent & Ambiguous        \\
			\hline
			SBU & 50.97\% & 12.88\% & 48.28\% & 20.85\% \\
			UCF & -       & -       & 60.91\% & 32.12\% \\
			\hline
		\end{tabular}
	}
	\label{tab:AmbiguousAdjacent}
\end{table}

\subsection{Statistical Comparison with State-of-the-art Shadow Detectors on Adjacent Shadows}
In order to present the performance of our proposed method on adjacent shadow detection more precisely, we calculate the proportion of adjacent shadows, including normal and ambiguous adjacent shadows, in SBU~\cite{vicente2016large} and UCF~\cite{zhu2010learning} datasets.
Table~\ref{tab:AmbiguousAdjacent} shows the statistical results on the two datasets, where ``Adjacent'' and ``Ambiguous'' refer to the percentages of adjacent shadows and ambiguous adjacent shadows in the whole dataset, respectively.
For SBU dataset, there are $50.97\%$ and $48.28\%$ images containing adjacent shadows in the training and testing set.
Furthermore, in the training and testing set, the ambiguous adjacent shadow reaches $12.88\%$ and $20.85\%$ of the whole dataset, respectively.
For UCF testing set, the percentage of adjacent shadow reaches $60.91\%$ with about one-third of the images in the whole test set are ambiguous adjacent shadows.
Figure~\ref{fig:adj} shows some cases of normal and ambiguous adjacent shadows we identified in the SBU dataset.
From the results, we can conclude that the adjacent shadows are ubiquitous in natural scenes, and many shadows are connected to the corresponding indistinguishable occluded objects.

\begin{table}[t]
	\renewcommand\arraystretch{1.05}  
	\caption{Resolved percentage (accuracy \%) comparison of our method with the state-of-the-art shadow detection methods on SBU~\cite{vicente2016large} dataset. ``Adjacent'', ``Normal'', and ``Ambiguous'' represent the percentage of shadow images successfully identified by the method in each category, respectively. Best results are highlighted in bold. ``$\uparrow$'' indicates the higher the better.}
	\centering
	\scalebox{0.98}{
		\begin{tabular}{
				m{2.2cm}<{\centering}
				|m{1.5cm}<{\centering}m{1.5cm}<{\centering}m{1.6cm}<{\centering}
			}
			\hline    
			& \multicolumn{3}{c}{SBU Dataset}   \\
			\hline
			methods & Adjacent$\uparrow$ & Normal$\uparrow$ & Ambiguous$\uparrow$   \\
			\hline
			ADNet~\cite{le2018a+}                     & 89.29\%  & 94.29\%  & 82.71\%    \\
			BDRAR~\cite{zhu2018bidirectional}         & 91.88\%  & 93.71\%  & 89.47\%    \\
			DSC~\cite{hu2018direction}                & 94.81\%  & 94.86\%  & 94.74\%    \\
			DSD~\cite{zheng2019distraction}           & 92.86\%  & 94.86\%  & 90.23\%    \\
			MTMT~\cite{chen2020multi}                 & 95.46\%  & 97.14\%  & 93.99\%    \\		
			FDRNet~\cite{zhu2021mitigating}           & 95.78\%  & 97.14\%  & 93.40\%    \\
			\hline
			\textbf{Ours}   & \textbf{97.08\%} & \textbf{97.72\%} & \textbf{96.24\%}     \\    
			\hline
		\end{tabular}
	}
	\label{tab:sbu_per}
\end{table}

\begin{table}[t]
	\renewcommand\arraystretch{1.05}  
	\caption{Resolved percentage (accuracy \%) comparison of our method with the state-of-the-art shadow detection methods on UCF~\cite{zhu2010learning} dataset. ``Adjacent'', ``Normal'', and ``Ambiguous'' represent the percentage of shadow images successfully identified by the method in each category, respectively. Best results are highlighted in bold. ``$\uparrow$'' indicates the higher the better.}
	\centering
	\scalebox{0.98}{
		\begin{tabular}{
				m{2.2cm}<{\centering}
				|m{1.5cm}<{\centering}m{1.5cm}<{\centering}m{1.5cm}<{\centering}
			}
			\hline    
			& \multicolumn{3}{c}{UCF Dataset}   \\
			\hline
			methods & Adjacent$\uparrow$ & Normal$\uparrow$ & Ambiguous$\uparrow$   \\
			\hline
			ADNet~\cite{le2018a+}                     & 70.15\%  & 90.32\%  & 52.78\%    \\
			DSC~\cite{hu2018direction}                & 82.09\%  & 90.32\%  & 75.00\%    \\
			DSD~\cite{zheng2019distraction}           & 77.61\%  & 90.32\%  & 66.67\%    \\
			MTMT~\cite{chen2020multi}                 & 83.58\%  & 90.32\%  & \textbf{77.78\%}    \\		
			FDRNet~\cite{zhu2021mitigating}           & 77.61\%  & \textbf{93.55\%}  & 63.89\%    \\
			\hline
			\textbf{Ours}   & \textbf{85.08\%} & \textbf{93.55\%} & \textbf{77.78\%}     \\    
			\hline
		\end{tabular}
	}
	\label{tab:ucf_per}
\end{table}

We compare our method with state-of-the-art shadow detection methods on SBU and UCF datasets, and make statistics on their ability to solve adjacent shadows.
The rules are as follows: 
We compare the predicted mask with the ground-truth shadow mask.
Given that there exist some noises, the predicted mask can not be completely consistent with ground-truth mask for every pixel.
So we stipulate that when the accuracy rate of predicted mask pixels and ground-truth mask pixels reaches more than $90\%$, the prediction is correct.
As shown in Table \ref{tab:sbu_per} and Table \ref{tab:ucf_per}, the results show that our method outperforms other shadow detectors on SBU and UCF dataset.
More concretely, the proposed network achieves the best performance among adjacent shadows, normal adjacent shadows, and ambiguous adjacent shadows, indicating that our method is effective in adjacent shadow detection.

\begin{figure*}[t]
	\centering
	\includegraphics[width=0.98\linewidth]{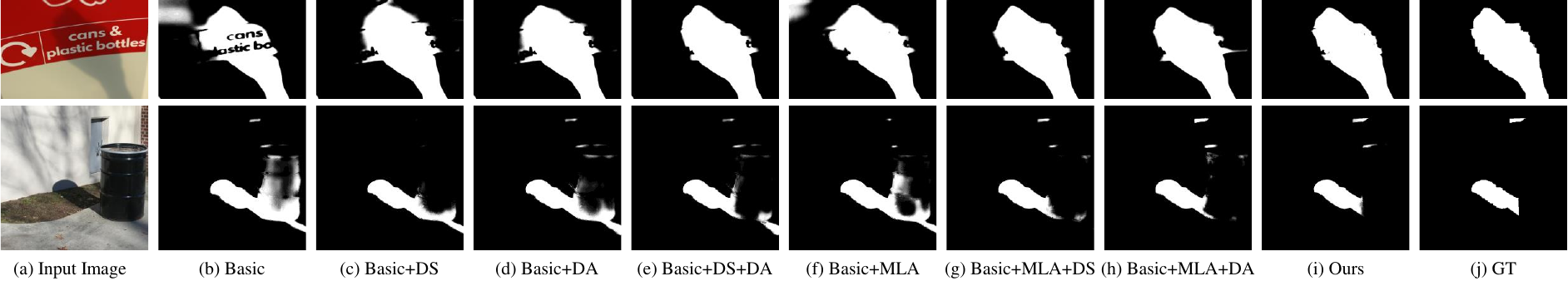}
	\caption{Visual comparison results of ablation studies.}
	\label{fig:ablation}
\end{figure*}

\subsection{Ablation Study on Model Components}
\label{exp:ablation}
In this section, we perform ablation studies to verify the effectiveness of each component of our network, and all experiments are conducted on the SBU dataset~\cite{vicente2016large}. 
Here we consider seven baseline networks.

The first four baseline networks are designed by removing the multi-level feature aggregation (MLA) mechanism and replaced with fusion modules that other models use~\cite{hou2017deeply}.
Specifically, the first baseline (denoted as "basic") only considers the transformer encoder backbone. 
The second baseline (denoted as "basic+DS") considers the deep supervision (DS) module, and the third baseline (denoted as "basic+DA") considers the double attention (DA) module. 
In contrast, the fourth baseline (denoted as "basic+DS+DA") considers both the DS and DA modules.

Another three baseline networks are designed by leveraging the proposed MLA mechanism. 
Specifically, the first one (denoted as "basic+MLA") considers the transformer encoder combined with the MLA module.
The second one (denoted as "basic+MLA+DS") considers the DS module, while the third one (denoted as "basic+MLA+DA") considers the DA module but ignores the DS module.

Table \ref{tab:ablation} reports the quantitative comparison results, representing that any module of DS, DA, and MLA can improve the shadow detection results.
Concretely, the DS module has a better effect than the other two.
We argue that the DS module can suppress the non-shadow region at the beginning of the model, enhance the shadow region according to the provided supervision signal, and transfer the enhanced features to all subsequent processes.
The DA module achieves a larger receptive field, integrating local and shifted window in one stage, which further boosts the relevance of the local signs and provides a flexible attention space for the network.
For the results using MLA, we conclude that the improved fusion mechanism decreases the difficulty of fusing features of different scales, resulting that each scale has the ability to localize the shadow regions.

Figure \ref{fig:ablation} shows the visual comparisons provided by ours and seven baseline networks.
Adding DS or DA module to the network improves both shadow prediction quality and localization accuracy.
Moreover, considering all components can lead to remarkable visual results, which proves the effectiveness and reliability of our methods.

\setlength{\tabcolsep}{7pt}
\begin{table}[t]
	\renewcommand\arraystretch{1.05}  
	\caption{Ablative experiments results on model components. DS, DA, and MLA stand for deep supervision module, double attention module, and multi-level aggregation mechanism, respectively. ``BER'' represents the balance error rate considering both shadow and non-shadow regions. Best results are highlighted in bold. ``$\downarrow$'' indicates the lower the better.}
	\small
	\centering
	\begin{tabular}{c|c|c|c|c}
		\hline
		& DS & DA & MLA & BER$\downarrow$ \\
		\hline
		basic          & $\times$     & $\times$     & $\times$     & 3.41 \\
		basic+DS       & $\checkmark$ & $\times$     & $\times$     & 3.04 \\
		basic+DA       & $\times$     & $\checkmark$ & $\times$     & 3.14 \\
		basic+DS+DA    & $\checkmark$ & $\checkmark$ & $\times$     & 2.92 \\
		\hline
		basic+MLA      & $\times$     & $\times$     & $\checkmark$ & 3.16 \\
		basic+MLA+DS   & $\checkmark$ & $\times$     & $\checkmark$ & 2.89 \\
		basic+MLA+DA   & $\times$     & $\checkmark$ & $\checkmark$ & 2.96 \\
		\hline
		Ours           & $\checkmark$ & $\checkmark$ & $\checkmark$ & \textbf{2.78} \\
		\hline
	\end{tabular}
	\label{tab:ablation}
\end{table}

\setlength{\tabcolsep}{15pt}
\begin{table}[t]
	\renewcommand\arraystretch{1.15}  
	\caption{Ablative experiments results on the backbone network. ``BER'' represents the balance error rate considering both shadow and non-shadow regions. Best results are highlighted in bold. ``$\downarrow$'' indicates the lower the better.}
	\small
	\centering
	\begin{tabular}{c|ccc}
		\hline
		& \multicolumn{3}{c}{BER$\downarrow$} \\
		\hline
		methods & SBU & UCF & ISTD    \\
		\hline
		ResNeXt-101~\cite{xie2017aggregated}     & 3.48 & 6.82 & 1.46  \\
		ViT-B~\cite{dosovitskiy2020image}           & 4.77 & 8.21 & 1.88  \\
		Swin-B~\cite{liu2021swin}    & \textbf{2.78} & \textbf{5.99} & \textbf{1.22}  \\
		\hline
	\end{tabular}
	\label{tab:res_vit_swin}
\end{table}

\begin{figure}[t]
	\centering
	\includegraphics[width=0.9\linewidth]{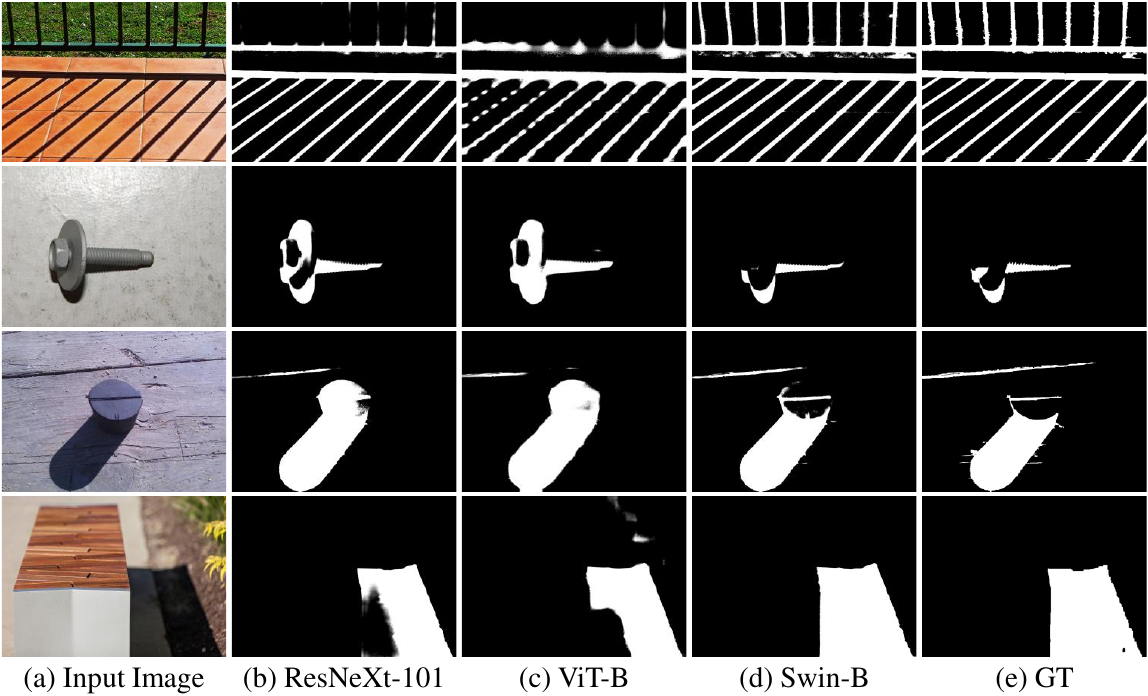}
	\caption{Visual comparison results of different backbone networks.}
	\label{fig:res_vit_swin}
\end{figure}

\subsection{Ablation Study on the Backbone Network}

Swin Transformer~\cite{liu2021swin} combines a self-attention mechanism with a hierarchical architecture.
In contrast, most CNN-based backbones like ResNeXt-101~\cite{xie2017aggregated} primarily feature a hierarchical structure alone, while Vision Transformer~\cite{dosovitskiy2020image} exclusively utilizes a self-attention mechanism.
Additionally, Swin Transformer~\cite{liu2021swin} introduces a shifted window that enhances interactions between windows, thereby improving its ability to distinguish shadows and black objects.
Its hierarchical structure is also well-suited for building multi-stage networks for pixel-level dense prediction tasks.
To further validate the advantages of Swin Transformer~\cite{liu2021swin} over CNN and Vision Transformer~\cite{dosovitskiy2020image}, we substitute our backbone network with ResNeXt-101~\cite{xie2017aggregated} and ViT-B~\cite{dosovitskiy2020image} for a comparative experiment.
The quantitative results, presented in Table~\ref{tab:res_vit_swin},  indicate that our method based on the Swin-B~\cite{liu2021swin} outperforms others.
Moreover, as shown in Figure~\ref{fig:res_vit_swin}, we provide qualitative results to further validate the effectiveness of the shifted window mechanism in Swin Transformer~\cite{liu2021swin}.

\begin{figure}[t]
	\centering
	\includegraphics[width=0.9\linewidth]{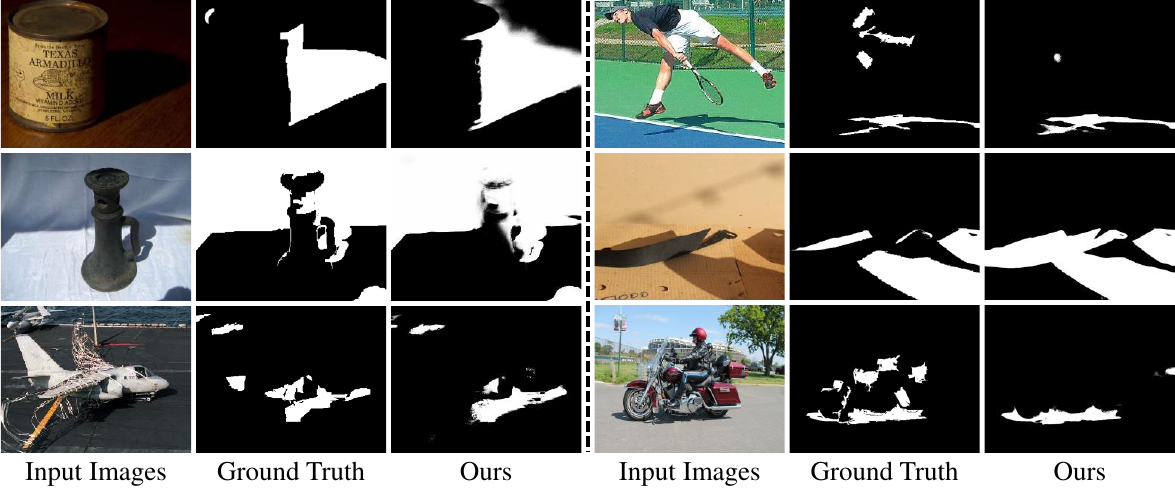}
	\caption{Failure cases.}
	\label{fig:failure}
\end{figure}

\subsection{Discussion}
\label{exp:discuss}
%

\textbf{Failure cases.}
We present more failure cases in Figure~\ref{fig:failure}. 
In this paper, our method fails in some complex cases where shadows overlap with black things, which can be categorized into two types: one where the shadow overlaps with a black background, and the other where it overlaps with black objects.
As shown in the three examples in the left column of Figure~\ref{fig:failure}, the appearance of the shadows and backgrounds is very similar, complicating the detection of shadow.
While for the three examples in the right column, there is partial overlap between the shadows and black objects.
Both situations pose challenges for the network.
We analyze that with the shifted window mechanism, the model can access more information. 
However, once shadows overlap with other black areas, the network struggles to determine the attributes of the overlapped portions. 
For the non-overlapping sections, the uncertainty of the overlapped parts prevents the model from making a semantic judgement about the entire shadow or object, leading to incorrect network predictions.
Hence, it is hard to distinguish even if we pay the increased attention to the entire image.
One possible solution is to weaken the influence of the background so that it can be transformed into the problem of adjacent shadows, which we propose a pipeline to resolve in this paper.

\section{Conclusion}
\label{conclusion}
This paper presents a transformer-based pipeline SwinShadow by introducing the powerful self-attention mechanism of the transformer for shadow detection, especially adjacent shadow detection.
The key idea of our method is to combine the local attention and shifted window mechanism.
Specifically, we propose a DS module to extract features from images, which can further enhance global feature extraction: highlight shadow features and suppress non-shadow features.
In addition, we use an effective MLA fusion method to interact with multi-scale features, transfer the shadow detail locality and high-level semantic information, and then improve the prediction accuracy of each score map.
In this way, the network can improve the identifiability of the object and its shadow.
Moreover, for the cases that shadows around a black object (see Figure \ref{fig:qualitative} last row), the model can also detect the shadows well. 
Compared with various latest methods, our method demonstrates promising performance in terms of balance error rate (BER) on three benchmark datasets.

As described in the paper, our method can address the adjacent shadows well, making the proposed method more applicable.
However, as shown in Figure \ref{fig:failure}, the model will be disturbed by the weak dark background. 
In future work, we will consider how to separate the background in the image to achieve a more accurate shadow detection effect.

\begin{acks}
This work was supported by National Natural Science Foundation of China under Contract 62021001, and the Youth Innovation Promotion Association CAS. 
It was also supported by GPU cluster built by MCC Lab of Information Science and Technology Institution, USTC, and the Supercomputing Center of the USTC.
\end{acks}

\bibliographystyle{ACM-Reference-Format}
\bibliography{sample-base}


\end{document}